\documentclass[10pt,twocolumn,letterpaper]{article}

\usepackage[pagenumbers]{cvpr} % To force page numbers, e.g. for an arXiv version

\usepackage{graphicx}
\usepackage{amsmath}
\usepackage{amssymb}
\usepackage{booktabs}
\usepackage[accsupp]{axessibility}

\usepackage[pagebackref,breaklinks,colorlinks]{hyperref}

\usepackage[capitalize]{cleveref}
\crefname{section}{Sec.}{Secs.}
\Crefname{section}{Section}{Sections}
\Crefname{table}{Table}{Tables}
\crefname{table}{Tab.}{Tabs.}

\def\cvprPaperID{6} % *** Enter the CVPR Paper ID here
\def\confName{CVPR}
\def\confYear{2023}

\begin{document}

\title{Self-supervised 3D Human Pose Estimation from a Single Image}

\author{Jose Sosa, David Hogg \\
School of Computing, University of Leeds\\
{\tt\small \{scjasm, D.C.Hogg\}@leeds.ac.uk}
}
\maketitle

%%%%%%%%% ABSTRACT
\begin{abstract}

We propose a new self-supervised method for predicting 3D human body pose from a single image. The prediction network is trained from a dataset of unlabelled images depicting people in typical poses and a set of unpaired 2D poses. By minimising the need for annotated data, the method has the potential for rapid application to pose estimation of other articulated structures (e.g. animals). The self-supervision comes from an earlier idea exploiting consistency between predicted pose under 3D rotation. Our method is a substantial advance on state-of-the-art self-supervised methods in training a mapping directly from images, without limb articulation constraints or any 3D empirical pose prior. We compare performance with state-of-the-art self-supervised methods using benchmark datasets that provide images and ground-truth 3D pose (Human3.6M, MPI-INF-3DHP). Despite the reduced requirement for annotated data, we show that the method outperforms on Human3.6M and matches performance on MPI-INF-3DHP. Qualitative results on a dataset of human hands show the potential for rapidly learning to predict 3D pose for articulated structures other than the human body. Project page: \url{https://josesosajs.github.io/imagepose/}

\end{abstract}

%%%%%%%%% BODY TEXT
\section{Introduction}
\label{sec:intro}

Estimating 3D pose for articulated objects is a long-standing problem. Its foundations arise from the early days of computer vision with model-based approaches representing the human body as an articulated structure of parts \cite{nevatia:1977, hogg1983model}. Interest in estimating 3D human pose grew within the computer vision community motivated by the many real-world applications, for example, pedestrian detection \cite{kim2019pedx}, human-computer interaction \cite{zhang2012microsoft}, video surveillance \cite{zheng2019pose}, and sports analysis \cite{rematas2018soccer}.

\begin{figure}[ht]
\centering
\includegraphics[width=0.99\linewidth]{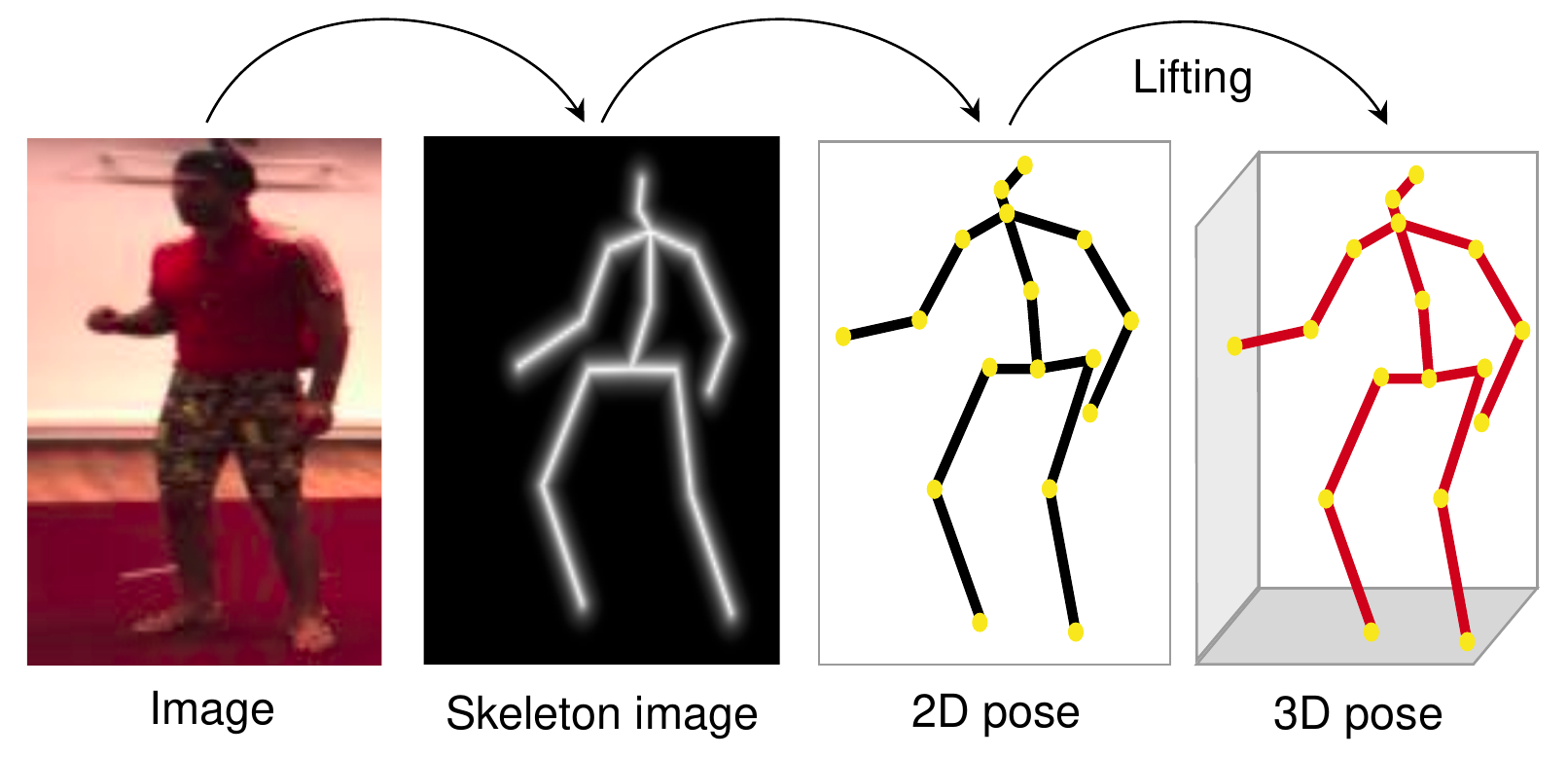}
\caption{\textbf{3D pose estimation pipeline.} Our approach jointly learns to estimate 3D pose from an image via intermediate representations of 2D pose. The pipeline is embedded within a larger network for end-to-end training. } 
\label{fig:summary-model}
\end{figure}

Initial work on estimating 3D pose addressed this problem by extracting a set of hand-crafted features, for example, segmentation masks \cite{agarwal2005recovering}. Other early approaches, such as exemplar-based methods, use extensive datasets of 3D poses (commonly constructed from motion capture data) to search for the optimal 3D pose given its 2D projection\cite{jiang20103d, gupta20143d, chen20173d}. Subsequently, these approaches were surpassed in performance by deep learning methods, initially using supervised learning to regress from images to joint positions \cite{toshev2014deeppose} or heatmaps \cite{tompson2015efficient, newell2016stacked}. While annotations for 2D joint positions in the image plane are relatively easy to obtain, getting ground truth 3D joint positions from images alone is not straightforward. 

Because of the limited availability of 2D and 3D pose annotations, an immense amount of available data in the form of images remains unexploited for training in pose estimation. Although there are many annotated datasets for human pose estimation, the situation is very different for non-human articulated structures such as animals. Recently, unsupervised and self-supervised methods for 3D human pose estimation have made progress using this unlabelled data (at least lacking 3D annotations) and have demonstrated that it is possible to learn to estimate 3D poses from 2D poses relying on rotational consistency \cite{chen2019unsupervised, drover2018can, yu2021towards}, multi-view setup \cite{iqbal2020weakly,rhodin2018unsupervised}, and 3D structure constraints e.g. joint angles between limbs \cite{kundu2020kinematic}. However, what will happen if we do not assume the availability of paired 2D poses? Is it still possible to train a model to predict 3D poses?  What are the minimum assumptions to make this possible?

We have proposed a method which, for the first time, learns to map between images and 3D pose without requiring 3D pose annotations or paired 2D pose annotations. Training only needs a set of unlabelled images depicting people in different poses and an unrelated set of 2D human poses. The motivation is twofold: (1) there is the potential for exceeding current levels of performance by training on massive unlabelled datasets, and (2) the method could, in principle, be applied to articulated structures (e.g. animals) where little or no 2D/3D annotated data is available. In our proposed method, we learn a 2D pose predictor and a 3D `lifting' function to produce 3D joint positions from unlabelled images (summarised in \autoref{fig:summary-model}) in an end-to-end learning framework. 

Our method simultaneously learns 2D and 3D pose representations in a largely unsupervised fashion, requiring only an empirical prior on unpaired 2D poses. We demonstrate its effectiveness on Human3.6M \cite{ionescu2013human3} and MPI-INF-3DHP \cite{mono-3dhp2017} datasets, two of the most popular benchmarks for human pose estimation. We also show the method's adaptability to other articulated structures using a synthetic dataset of human hands \cite{simon2017hand}. In experiments, the approach outperforms state-of-the-art self-supervised methods that estimate 3D pose from images and require higher supervision in training. Overall, our method has the following advantages:

\begin{itemize}
    \item It does not assume any 3D pose annotations or paired 2D pose annotations.
    \item It holds the potential for quickly adapting to 3D pose prediction for other articulated structures (e.g. animals and jointed inanimate objects).
\end{itemize}

\section{Related Work}
\label{sec:relwork}

Our method broadly relates to prior work that estimates 3D human pose directly from images, and mainly to self-supervised deep learning methods. However, it also draws inspiration from earlier work on the estimation of 3D pose from 2D pose. Therefore, we review both perspectives, regardless of the degree of supervision required for training.

\subsection*{3D pose from 2D pose}
A range of methods take as input 2D poses and lift them to 3D space. Frequently, the 2D poses come from an off-the-shelf 2D pose estimator, or they are simply annotations for a given dataset. Early techniques for estimating 3D poses from 2D joint positions rely on classical classification algorithms and physical constraints. For example, given the joint connectivity and bone lengths, \cite{lee1985determination} use binary decision trees to estimate the two possible states of a joint with respect to its parent. Other methods implement the nearest neighbour algorithm with large datasets of 3D poses to search for the most likely 3D representation of a particular 2D pose \cite{gupta20143d, chen20173d, jiang20103d}.

In the deep learning context, Martinez \etal\cite{martinez2017simple} present a fully supervised approach to predict 3D positions given 2D joint locations using a fully connected network with residual blocks. This network structure has become popular, and subsequent unsupervised approaches \cite{drover2018can, chen2019unsupervised, yu2021towards, wandt2021elepose} incorporate it within their processes. For instance, Drover \etal \cite{drover2018can} propose a weakly supervised approach that lifts a 2D pose to 3D and then evaluates its 2D projection through a GAN loss. Later, Chen \etal \cite{chen2019unsupervised} extended this work by adding a symmetrical pipeline of consecutive transformations (lifting, rotation, and projection) of the estimated 3D representation. This cycle of transformations exploits geometric consistency and removes the dependency on any 3D correspondences. 

More recently, Wandt \etal\cite{wandt2021elepose} incorporate two fundamental elements to the model in \cite{chen2019unsupervised} that increase the performance of the 3D lifting process: the use of normalising flow (NF) and a learned elevation angle for the 3D rotations. Previous methods have successfully used normalising flow to estimate 3D prior distributions given 3D human poses \cite{wehrbein2021probabilistic}. However, the method in \cite{wandt2021elepose} is the first to use normalising flow to infer the probability of a reconstructed 3D pose from a prior distribution of the 2D input.  

Unlike these previous methods, we do not assume access to ground truth 2D poses as input. Instead, our model takes a single image and predicts the 2D pose from it, which is then lifted to 3D. Overall, it estimates both the 2D and 3D poses from the input image, removing the dependency on paired 2D pose annotations or pre-trained 2D pose predictors.

\begin{figure*}[!ht]
\centering
\includegraphics[width=\textwidth]{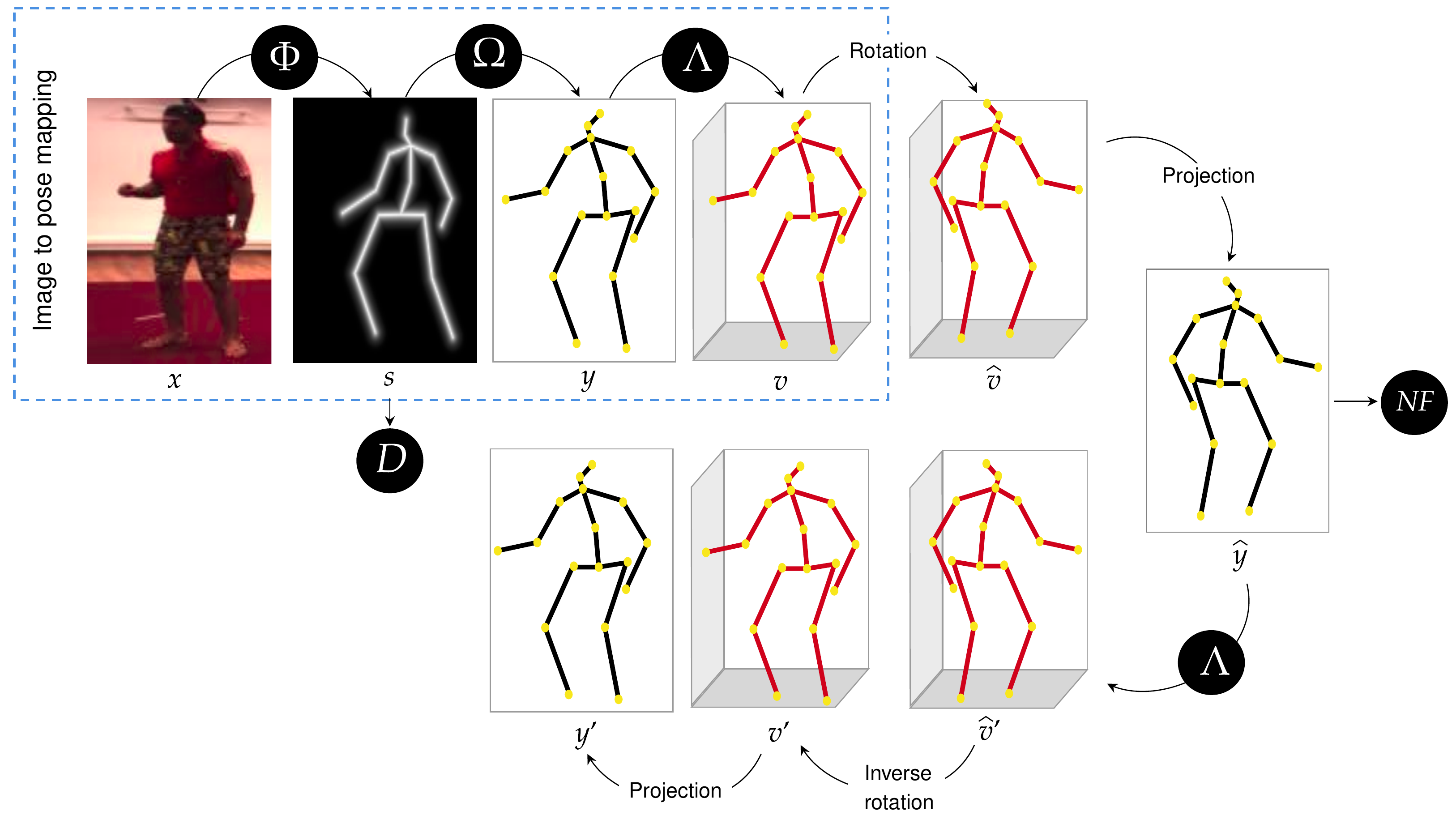}
\caption{\textbf{Self-supervised architecture for estimating the 3D pose of a person.}  Our method aims to map an image of a person $x$ to its 3D pose $v$. To achieve this; we train the networks $\Phi, \Omega, D,$ and $\Lambda$ with help of a normalising flow network and a couple of rotations and projections. The pipeline starts with passing an input  image $x$ through $\Phi$ to obtain a skeleton image $s$, then $\Omega$ produces a set of 2D joint positions $y$ given $s$. We then  embed the mapping into a larger cyclic architecture, where the 2D pose $y$ is lifted into 3D using $\Lambda$. The 3D pose $v$ is then rotated, $\hat{v}$ projected into 2D,  $\hat{y}$ is lifted back again into 3D through the same $\Lambda$, $\hat{v}'$ inversely rotated, and finally $v'$ projected back to the 2D pose $y'$ which should resemble the original 2D pose $y$.}
\label{fig:whole-model}
\end{figure*}

\subsection*{3D pose from an image}
Work under this category is more related to our approach. Typically, methods for estimating 3D pose from single images break down the task into two steps. First, the 2D joints are localised, and then the 3D pose is estimated from these 2D joint positions. Early deep-learning implementations of this two-step process depend on having access to 2D ground truth joint locations and 3D data for supervising training  \cite{li20143d, park20163d, zhou2017towards}. Furthermore, most of those methods integrate pre-trained 2D pose estimators, e.g. a stacked hourglass network \cite{newell2016stacked} and AlphaPose \cite{fang2017rmpe}, for solving the joint localization step.

On the unsupervised side, many approaches incorporate specific assumptions for the 2D and 3D joint configuration or even add some small portion of actual 3D data to guide the training. For instance, \cite{tome2017lifting} shows a unified multi-stage CNN architecture to estimate 2D and 3D joint locations from single images. This approach relies on a probabilistic 3D model of the human pose responsible for lifting the 2D representations. Other methods incorporate motion information; for example, Zhou \etal\cite{zhou2016sparseness} use sequences of images and their corresponding 2D pose to guide their 3D pose estimation framework. 

Like our work, Kundu \etal\cite{kundu2020self} propose a self-supervised architecture to learn 3D poses from unlabelled images. They incorporate three assumptions: human pose articulation constraints, a part-based 2D human puppet model, and unpaired 3D poses. Other approaches explore learning without direct supervision by producing synthetic multi-views of the same skeleton \cite{rhodin2018unsupervised}, accessing multi-view videos \cite{mitra2020multiview}, or relying solely on 3D kinematic constraints \cite{kundu2020kinematic}. In contrast, our method does not depend on multi-view images nor any 3D data; it requires only an unlabelled dataset of images depicting people in typical poses and an unpaired empirical prior for the 2D pose.

%-------------------------------------------------------------------------
\section{Method}

Our proposed 3D pose estimation model consists of a pipeline of three networks $\Phi, \Omega, \Lambda$ mapping from full body images to 3D pose. This is shown in the upper-left blue dotted box in \autoref{fig:whole-model}. The pipeline consists of:
\begin{itemize}
    \item a Convolutional Neural Network (CNN) $\Phi$ mapping from an input image $x$ to an intermediate skeleton image $s$,
    \item a second CNN $\Omega$ mapping from $s$ to a 2D pose representation $y$, and
    \item a fully connected network $\Lambda$ lifting the 2D pose $y$ to the required 3D pose $v$.
\end{itemize}
The 3D pose is represented as an articulated structure of 3D line segments corresponding to the parts of the body (e.g., torso, head, upper arm, foot).

We train the three networks together by incorporating into a larger network (\autoref{fig:whole-model}) and optimise end-to-end. This larger network is constructed to incorporate a loop of transformations of the 3D pose. The degree of geometric consistency around the loop contributes to a loss function and provides self-supervision of the training. The training starts with a dataset of images depicting people in different poses. We also assume we have a (normally unrelated) dataset of typical 2D poses from which we obtain skeleton images using a differentiable rendering function $\kappa$. These will be used in a GAN framework $D$ to help ensure the generated skeleton images are realistic. In the following sections, we provide more details about the components of our model.

\subsubsection*{Image to 3D pose mapping}
The image-to-pose mapping is the composition of networks $\Phi,\Omega, \Lambda$ to map an image $x$ showing a person to its 3D pose representation $v$. The first part of the mapping is  a CNN $\Phi$, which maps from the image $x$ to a skeleton image $s = \Phi(x)$ showing the person as a stick figure. Our network $\Phi$ adopts a similar architecture to the autoencoder in \cite{jakab:2020} but without the decoder stage. After training, the emergent skeleton in $s$ aligns with the person in $x$ as expected.

Then, the network $\Omega$ maps the skeleton image $s$ to a 2D pose representation $y = \Omega(\Phi(x))$. Informally, $\Omega$ learns to extract 2D joint positions $(x_i, y_i)$ from the skeleton image. Finally, $\Lambda$ is a neural network that lifts the 2D pose to the required pose $v$ in 3D.  In particular,  $\Lambda(y)$ estimates the depth $z_i = d_i + \Delta$ for each pair of $(x_i,y_i)$ joint positions in the input $y$, where $\Delta$ is a constant depth. Then, the 3D position of joint $v_i$ in the 3D pose $v$ is given by

\begin{equation}
    v_i = (x_i \cdot z_i, y_i \cdot z_i, z_i) 
\end{equation}

where $z_i$ is forced to be larger than one, to prevent ambiguity from negative depths. In line with previous works \cite{chen2019unsupervised, yu2021towards, wandt2021elepose}, $\Delta$ is fixed to $10$.

Our lifting network $\Lambda$ is based on the work of \cite{martinez2017simple, chen2019unsupervised} and extended following \cite{wandt2021elepose}. In this context, our extended version not only outputs the depth $z_i$ for each joint position $(x_i, y_i)$ in the input, it also produces a value for the elevation angle $\alpha$. This angle will be used when performing the rotations of the 3D pose $v$. In particular, we use $\alpha$ to fix the elevation angle of the vertical axis to the ground-plane about which the rotation is performed. 

\subsubsection*{Skeleton images and discriminator}
\label{discrminator:skeleton image}

We encourage the training network to generate realistic skeleton images with the help of an empirical prior of 2D poses. Note that these 2D poses are unpaired, i.e., they are not annotations of the training images. 

The 2D poses from our empirical prior are first rendered as skeleton images using the renderer proposed by \cite{jakab:2020}. Let $C$ be a set of connected joint pairs $(i,j)$, $e$ an image pixel location, and $u$ a set of $(x,y)$ 2D coordinates of body joint positions. The skeleton image rendering function is given by:
\begin{equation}\label{eq:render}
    \kappa(u)_{e} = \exp \biggl( -\gamma \min\limits_{(i,j) \in C, r \in [0,1]} || e - ru_{i} - (1 - r)u_{j} || ^ 2 \biggr)
\end{equation}

Informally $\kappa$ works by defining a distance field from the line segments linking joints and applies an exponential fall-off to create the image.

Following \cite{jakab:2020}, we use a discriminator network $D$ that uses the prior skeleton images to encourage the predicted skeleton images to represent plausible poses. The task of $D$ is to determine whether or not a skeleton image $s = \Phi(x)$ looks like an authentic skeleton image such as those in the prior $w = \kappa(u)$. Formally, the goal is to learn $D(s) \in [0,1]$ to match the reference distribution of skeleton images $p(w)$ and the distribution of predicted skeleton images $q(s)$. An adversarial loss \cite{NIPS2014_5ca3e9b1} compares the unpaired samples $w$ and the predictions $s$:

\begin{equation} \label{eq:disc}
    \mathcal{L}_D = \mathbb{E}_w(log(D(w)) + \mathbb{E}_s(log(1 - D(s))
\end{equation}

\subsubsection*{Random rotations and projections}
A fundamental component of our model is the lifting process which allows learning an accurate 3D pose $v$ from the estimated 2D input $y$. To provide self-supervision of the lifting function and ultimately the whole end-to-end network, we emulate a second virtual view of the 3D pose $v$ by randomly rotating it $\hat{v} = R * v$ . Previous work \cite{chen2019unsupervised} has selected a rotation matrix $R$ by uniformly sampling azimuth and elevation angles from a fixed distribution. Recently, \cite{wandt2021elepose} demonstrates that learning the distribution of elevation angles leads to better results. Thus, we follow their approach and use $\Lambda$ to also predict the elevation angle for the rotation matrix. The rotation around the azimuth axis $R_a$ is sampled from a uniform distribution $[-\pi, \pi]$.

In line with \cite{wandt2021elepose}, we also predict the dataset's normal distribution of elevation angles $R_e$ by calculating a batchwise mean $\mu_e$ and standard deviation $\sigma_e$. We sample from the normal distribution $\mathcal{N}(\mu_e, \sigma_e)$ to rotate the pose in the elevation direction $R_e$. Then, the complete rotation matrix $R$ is given by $R = R^{T}_{e}R_aR_e$.

After rotating the 3D pose, we project $\hat{v}$ through a perspective projection. Then, the same lifting network $\Lambda(\hat{y})$ produces another 3D pose $\hat{v}'$ which is then rotated back to the original view. The final 3D pose $v'$ is projected to 2D using the same perspective projection. This loop of transformations of the 3D pose provides a self-supervised consistency loss. In this context, we assume that if the lifting network $\Lambda$ accurately estimates the depth for the 2D input $y$, then the 3D poses $\hat{v}$ and $\hat{v}'$ should be similar. The same principle applies to $y$ and the final 2D projection $y'$. This gives the following two components of the loss function:

\begin{equation}
\label{eq:l3}
    \mathcal{L}_{3d} = || \hat{v}' -  \hat{v} ||^2
\end{equation}

\begin{equation}
\label{eq:l2}
    \mathcal{L}_{2d} = || y' - y ||^2
\end{equation}

Additionally, the 3D poses $v$ and $v'$ should be similar. However, instead of comparing with an $L_2$ loss, we follow \cite{yu2021towards,wandt2021elepose} and measure the change in the difference in 3D pose between two samples from a batch at corresponding stages in the network. 

\begin{equation}
\label{eq:def}
    \mathcal{L}_{def} = || ( {v'}^{(j)} - {v'}^{(k)} ) - ( {v}^{(j)} - {v}^{(k)} ) || ^ 2
\end{equation}

Similar to Wandt \etal\cite{wandt2021elepose}, we do not assume samples are from the same video sequence; the samples $j$ and $k$ may come from different sequences and subjects.

\subsubsection*{Empirical prior on 2D pose}

Like Wandt \etal\cite{wandt2021elepose}, we use a normalizing flow to provide a prior over 2D pose. A normalising flow transforms a simple distribution (e.g. a normal distribution) into a complex distribution in such a way that the density of a sample under this complex distribution can be easily computed. 

Let $Z \in \mathbb{R}^N$ be a normal distribution and $g$ an invertible function $g(z) = \bar{y}$ with $\bar{y} \in \mathbb{R}^N$ as a projection of the 2D human pose vector $\hat{y}$ in a PCA subspace. By a change of variables, the probability density function for $\bar{y}$ is given by

\begin{equation}
\label{eq:nf}
    p_Y(\bar{y}) = p_{Z}(f(\bar{y})) \biggl| \det\biggl( \frac{\delta f}{\delta \bar{y}} \biggr) \biggr|
\end{equation}

where $f$ is the inverse of $g$ and $\frac{\delta f}{\delta \bar{y}}$ is the Jacobian of $f$. Following the normalising flow implementation in \cite{wandt2021elepose}, we represent $f$ as a neural network \cite{dinh2016density} and optimise over a dataset of 2D poses with negative log likelihood loss:

\begin{equation}
\label{eq:nf_2}
    \mathcal{L}_{NF} = - \log(p_{Y}(\bar{y}))
\end{equation}

\subsubsection*{Additional losses}
We compute a loss from the mapping from skeleton images to 2D pose $y = \Omega(s)$. We use the same loss as \cite{jakab:2020}, but without pretraining $\Omega$, i.e., we learn this mapping simultaneously with all the other networks. $\mathcal{L}_{\Omega}$ is given by

\begin{equation}
\label{eq:reg}
    \mathcal{L}_{\Omega} =  || ( \Omega(\kappa(u)) -  u )||^2 + \lambda|| (\kappa(y) - s) ||^2
\end{equation}

where $u$ represents a 2D pose from the unpaired prior, $s$ is the predicted skeleton image, and $\lambda$ is a balancing coefficient set to $0.1$. The function $\kappa$ is the skeleton image renderer defined in \autoref{eq:render}.

Based on the proven effectiveness of incorporating relative bone lengths into pose estimation methods \cite{wandt2021elepose, nie2019view, li2019boosting}, we add this to impose a soft constraint when estimating the 3D pose. Following the formulation in \cite{wandt2021elepose}, we calculate the relative bone lengths $b_n$ for the n-th bone divided by the mean of all bones of a given pose $v$. We use a pre-calculated relative bone length $\bar{b}_n$ as the mean of a Gaussian prior. Then, the negative log-likelihood of the bone lengths defines a loss function $\mathcal{L}_{bl}$,

%\begin{equation}
%\label{eq:bone_len}
%    p(b_1, ... , b_N | \bar{p_1}, ... , \bar{p}_N) = \prod_{n = 1}^{N} \mu(b_n|\bar{b}_n, \sigma_b)
%\end{equation}

\begin{equation}
\label{eq:bl_sc}
    \mathcal{L}_{bl} = -\log (\prod_{n = 1}^{N} \mathcal{N}(b_n|\bar{b}_n, \sigma_b))
\end{equation}
%where N is the number of bones defined by the connectivity between joints. Then, the negative log-likelihood of \autoref{eq:bone_len} defines the loss function $\mathcal{L}_{bl}$,

where $N$ is the number of bones defined by the connectivity between joints. Note that this is a soft constraint that allows variation in the relative bone lengths between individuals. 

\subsection{Training}
We train $\Phi, \Omega, D,$ and $\Lambda$ from scratch. Only the normalising flow is independently pre-trained, as indicated in \cite{wandt2021elepose}. The complete loss function for training our model has seven components expressed as $\mathcal{L}_{D}$ (\autoref{eq:disc}), $\mathcal{L}_{\Omega}$ (\autoref{eq:reg}), $\mathcal{L}_{2d}$ (\autoref{eq:l2}), $\mathcal{L}_{3d}$ (\autoref{eq:l3}), $\mathcal{L}_{def}$ (\autoref{eq:def}),  $\mathcal{L}_{NF}$ (\autoref{eq:nf_2}), and $\mathcal{L}_{bl}$ (\autoref{eq:bl_sc}). For convenience in ablation studies, we group three of these loss terms and represent them as $\mathcal{L}_{base}$ 

\begin{equation}
\label{eq:base}
    \mathcal{L}_{base} = \mathcal{L}_{2d} + \mathcal{L}_{3d} + \mathcal{L}_{def}
\end{equation}

Thus, the final composite loss function is defined as:

\begin{equation}
\label{eq:completeloss}
    \mathcal{L} = \mathcal{L}_{D} + \mathcal{L}_{\Omega} + \mathcal{L}_{base} + \mathcal{L}_{NF} + \mathcal{L}_{bl}
\end{equation}

During testing we only keep the pipeline composed of the trained $\Phi$, $\Omega$, and $\Lambda$ networks illustrated in the upper-left box in \autoref{fig:whole-model}. Please see the supplementary section for a more detailed description of the networks and training.

\section{Experiments}
\subsection{Datasets}
\textbf{Human3.6M:} Human3.6M \cite{h36m:2014} is a widely used large-scale pose dataset consisting of videos of 11 subjects doing 17 activities against a static background. There are 3.6M frames depicting the human body and corresponding 3D body poses. In line with the standard protocol II on Human3.6M \cite{h36m:2014}, we use frames from subjects S1, S5, S6, S7, and S8 for training. For testing, we use frames from subjects S9 and S11. We pre-processed the video data by cropping the human body on each frame and removing the background, using the bounding boxes and segmentation masks provided in the dataset.

\textbf{MPI-INF-3DHP:} MPI-INF-3DHP \cite{mehta2017monocular} is a human pose dataset containing 3D annotations. Unlike Human3.6M, this dataset incorporates studio and outdoor recordings. The dataset comprises eight subjects with two video sequences for each, doing different activities, e.g. walking, sitting, exercising, and reaching. We train our model with the images from the train split and evaluate with the given test set. 

\textbf{HandDB:} HandDB\cite{simon2017hand} is a dataset of images showing human hands under different scenarios. For our experiments, we use part of the subset of hands generated from synthetic data, which contains 2D annotations for 21 key points: four for each of the five fingers and one for the wrist. For training and testing, we select two sequences (synth2 and synth3) of images from the four included in this subset and split them 80/20, respectively.

\subsection{Metrics}
Following previous methods \cite{martinez2017simple, chen2019unsupervised, drover2018can, rhodin2018unsupervised, mitra2020multiview}, we use the standard \textit{Protocol II} to evaluate the Human3.6M dataset \cite{h36m:2014}. \textit{Protocol II} performs a rigid alignment between the predicted 3D pose and the 3D ground truth via the Procrustes method \cite{goodall1991procrustes}. Then, it calculates the Mean Per Joint Position Error (MPJPE), which takes the average Euclidean distance between the ground-truth joint positions and the corresponding estimated positions across all 17 joints \cite{h36m:2014}. For simplicity, we refer to this metric as P-MPJPE (for Procrustes-MPJPE). For evaluation of the MPI-INF-3DHP dataset, we show the Percentage of Correct Keypoints (PCK), which represents the percentage of estimated joint positions within a fixed distance of $150mm$ from their respective ground truth. We also report the corresponding area under the curve (AUC).
%-------------------------------------------------------------------------
\begin{figure*}[!ht]
\centering
\includegraphics[width=\textwidth]{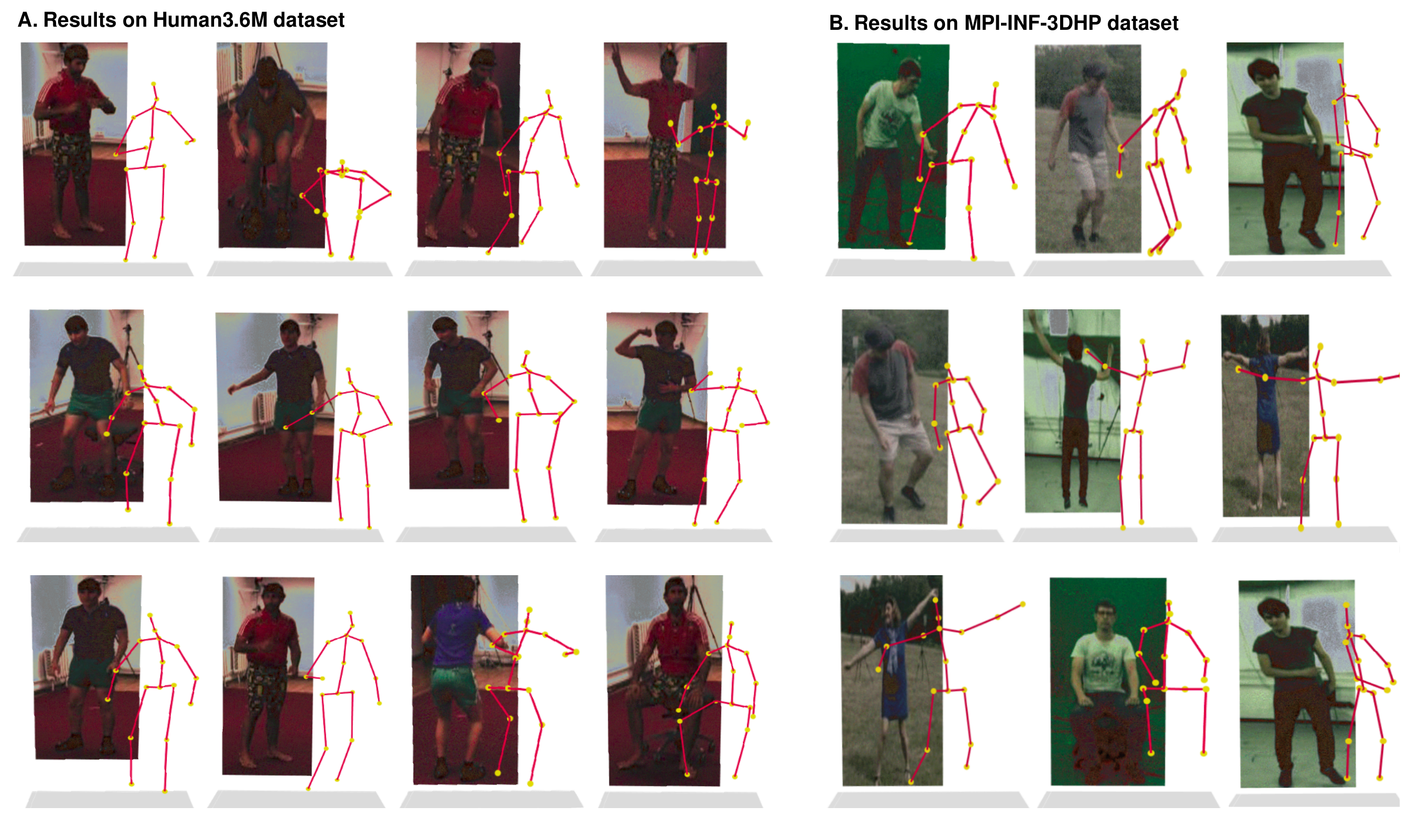}
\caption{\textbf{Qualitative results on images from Human3.6M and MPI-INF-3DHP datasets.} Each figure contains the input image and its corresponding estimated 3D pose by our model. An extended version of qualitative results for both datasets is included in the supplemental document.} 
\label{fig:vis-pred}
\end{figure*}

\subsection{Results}

Using our trained model with the Human3.6M dataset, we predict 3D pose (consisting of 17 joint positions) for every frame in all videos of subjects S9 and S11 ($\sim$ 584k frames) and calculate the average P-MPJPE. \autoref{table:comparison-table} compares our method with the state-of-the-art 3D pose estimation methods. 

\begin{table}[!ht]
\begin{center}
\begin{tabular}{|l|l|c|}
\hline
\textbf{Assumptions}    &   \textbf{Method}     &   \textbf{P-MPJPE($\downarrow$)}  \\
\hline 
\multicolumn{3}{|c|}{\textbf{3D pose from 2D landmarks}}\\
\hline
3D poses     &   Martinez \etal\cite{martinez2017simple}                 &   52.1\\
\hline 

%Full 2D     &   Sun \etal\cite{sun2021unsupervised}                     &   89.6 \\
2D poses     &   Chen \etal\cite{chen2019unsupervised}                   &   68.0 \\
2D poses     &   Drover \etal\cite{drover2018can}                        &   64.6\\
2D poses     &   Yu \etal\cite{yu2021towards}                            &   52.3 \\
2D poses     &   Wandt \etal\cite{wandt2021elepose}                      &   36.7 \\
\hline

\multicolumn{3}{|c|}{\textbf{3D pose from images}} \\
\hline
3D Poses     &   Chen \etal\cite{chen20173d}                             &   114.2 \\
3D Poses     &   Mitra \etal\cite{mitra2020multiview}                    &   72.5\\
\hline
MV + P3D      &   Rhodin \etal\cite{rhodin2018learning}                   &   128.6 \\
MV + P3D      &   Rhodin \etal\cite{rhodin2018unsupervised}               &   98.2 \\
MV + 2D      &   Wandt \etal\cite{wandt2021canonpose}                    &   53.0  \\
\hline
Unpaired 3D         &   Kundu \etal\cite{kundu2020self}                         &   99.2 \\
%NO3D        &   Kundu \etal\cite{kundu2020self}                         &   126.8 \\
3D Struct.        &   Kundu \etal\cite{kundu2020kinematic}                    &   89.4 \\
%NO3D        &   Kundu \etal\cite{kundu2020kinematic}                    &   134.8\\
\hline
Unpaired 2D        &   Ours        & 96.7\\
\hline
\end{tabular}
\end{center}
\caption{\textbf{P-MPJPE (in mm's) for all activities in Human3.6M.} MV = Multi view, Unpaired 3D = Unpaired 3D poses, 3D Struct. = 3D body structure constraints, Unpaired 2D = Unpaired 2D poses, P3D = Partial 3D poses (i.e, one portion of the 3D pose annotations available).}
\label{table:comparison-table}
\end{table}

We include supervised \cite{chen20173d, mitra2020multiview}, semi-supervised  \cite{rhodin2018learning, rhodin2018unsupervised, wandt2021canonpose}, and self-supervised \cite{kundu2020kinematic, kundu2020self} approaches that estimate the 3D pose from images. For a more comprehensive comparison, we also consider supervised \cite{martinez2017simple} and unsupervised \cite{sun2021unsupervised, chen2019unsupervised, drover2018can, yu2021towards, wandt2021elepose} methods that estimate 3D pose from 2D landmarks. The performance of our method exceeds that of some methods that rely on 3D supervision \cite{chen20173d}, multi-view images \cite{rhodin2018learning, rhodin2018unsupervised} or priors on 3D data \cite{kundu2020self}.

To demonstrate our model's generalisation performance, we evaluate using the MPI-INF-3DHP test dataset under different settings. First, we trained the model using Human3.6M. Second, we train our model with images from the MPI-INF-3DHP training set. Finally, we train with both images from the MPI-INF-3DHP training data and the training set of Human3.6M. For the first experiment, the set of 2D poses used for the normalising flow prior on 2D pose and the derived empirical prior on skeletons comes from Human3.6M, and for the rest the empirical prior comes from 2D poses on MPI-INF-3DHP. \autoref{table:mpi-inf} presents the PCK and AUC scores for the different evaluation conditions.

\begin{table}[!ht]
\begin{center}
\begin{tabular}{|l|l|c|c|}
\hline
\textbf{} & \textbf{Method} & \textbf{PCK($\uparrow$)} &  \textbf{AUC($\uparrow$)} \\
\hline
Unpaired 3D & Kundu \etal \cite{kundu2020self} & 83.2 & 58.7\\ 
3D Struct. & Kundu \etal \cite{kundu2020kinematic} & 79.2 & 43.4\\ 
\hline
Unpaired  2D  & Ours\textsuperscript{-}  &   58.7 &  24.3 \\
Unpaired  2D  & Ours\textsuperscript{*}  &   69.6 &  32.8 \\
Unpaired  2D  & Ours\textsuperscript{+}   &  75.3 & 40.0 \\ 
\hline
\end{tabular}
\end{center}
\caption{\textbf{Evaluation results on MPI-INF-3DHP dataset.} First column shows the main assumption for each method, where Unpaired 3D = Unpaired 3D poses, 3D Struct. = 3D body structure constraints, and Unpaired 2D = Unpaired 2D poses. Ours\textsuperscript{-} indicates the model trained with Human3.6M and tested with MPI-INF-3DHP. Ours\textsuperscript{*} represents the model trained with images from  MPI-INF-3DHP. Ours\textsuperscript{+} indicates that the MPI-INF-3DHP train set has been extended with images from Human3.6M.}
\label{table:mpi-inf}
\end{table}

\autoref{fig:vis-pred} shows qualitative results on images from Human3.6M and MPI-INF-3DHP datasets. It includes a random selection of predicted 3D poses and their corresponding input image.

\begin{figure}
\centering
\includegraphics[width=\linewidth]{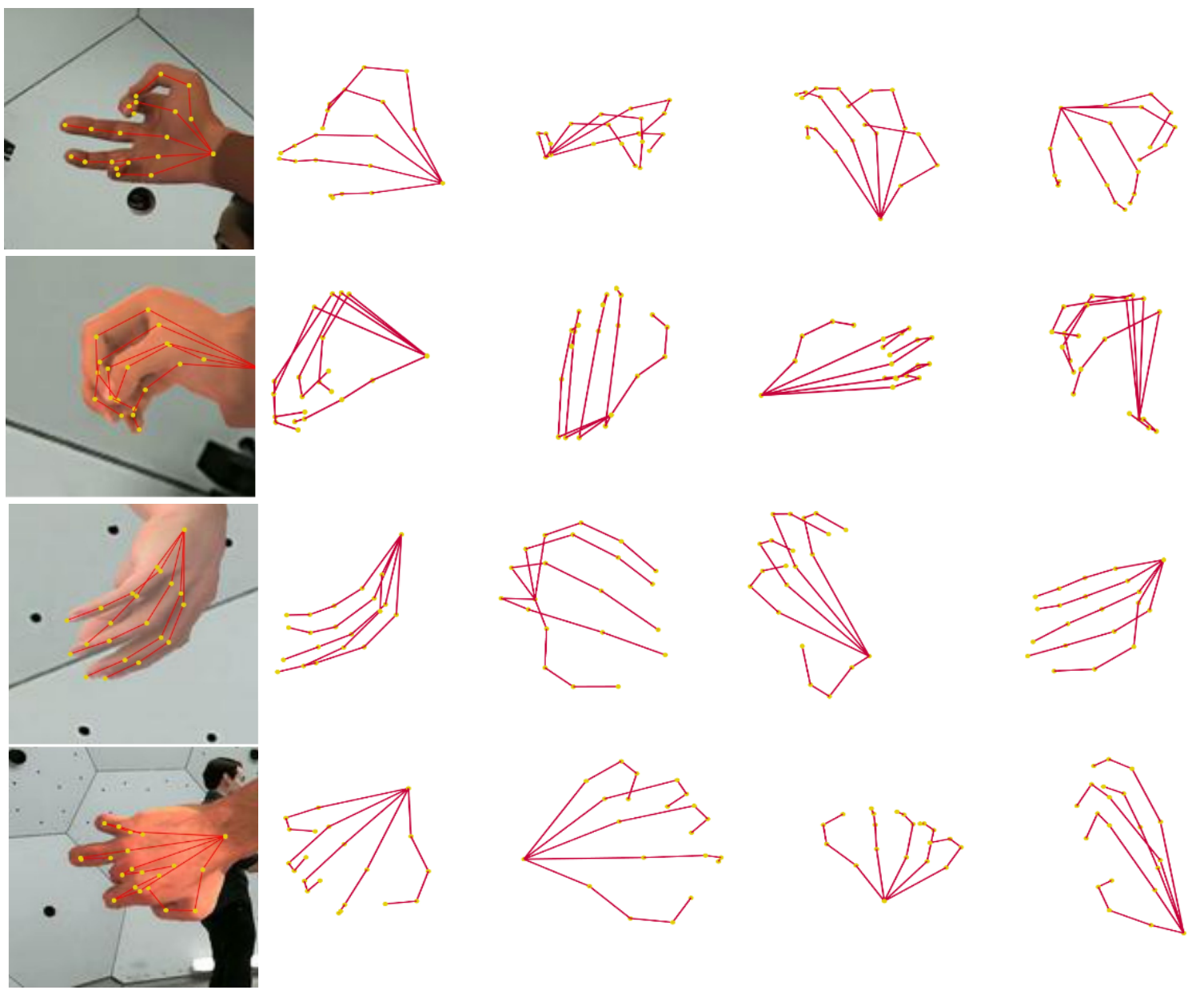}
\caption{\textbf{Qualitative results on HandDB dataset.} First column shows the input image overlay with the 2D ground truth annotations from the dataset. Next columns display novel views of the 3D hand predictions.} 
\label{fig:hands-vis}
\end{figure}

To demonstrate the adaptability of the method, we applied it to estimate hand pose using part of the synthetic subset from \cite{simon2017hand}. This required a different structure for the target 3D pose. For training and testing the model, we select sequences of images showing hands under similar conditions (synth2 and synth3). We augment the training set offline by making two rotated versions of each image ($45^{\circ}$ and $90^{\circ}$). We use half of the 2D annotations provided with the dataset to build the prior on 2D hand poses. The training set does not include the images corresponding to those annotations. With the trained model, we estimate 3D hand poses consisting of 21 key points (representing hand joint positions). Since the synthetic subset of HandDB does not contain 3D annotations (just 2D), we only show qualitative results in \autoref{fig:hands-vis}.

\subsection{Ablation study}
To evaluate the effectiveness of the design of our loss function, we progressively removed components from the complete loss expressed by \autoref{eq:completeloss}. First, we evaluate the model guided by the discriminator loss $\mathcal{L}_D$ and $\mathcal{L}_{\Omega}$. As expected, even when the 2D predictions are mostly accurate, the overall performance decreases since there is nothing else to regulate the 3D predictions and these are more likely to be deformed. We keep the $\mathcal{L}_D$, $\mathcal{L}_{\Omega}$, and $\mathcal{L}_{base}$ losses for the second experiment, i.e., removing $\mathcal{L}_{NF}$ and $\mathcal{L}_{bl}$. Although the model can produce plausible 3D poses, the performance is still inferior. 

Finally we assess the performance when removing only the loss term constraining the bone lengths $\mathcal{L}_{bl}$ from the original loss formulation (\autoref{eq:completeloss}). Adding the combination of the loss terms for the normalising flow prior on 2D pose $\mathcal{L}_{NF}$ and relative bone length $\mathcal{L}_{bl}$ has proven to be useful, increasing the performance of the model by $20.8\%$ with respect to the loss function that does not contain those terms. \autoref{table:ablation-table} shows the results for each of the modifications to the loss function. 

\begin{table}
\begin{center}
\begin{tabular}{|l|c|}
\hline
\textbf{Configuration} & \textbf{P-MPJPE($\downarrow$)}  \\
\hline
$\mathcal{L}_{D} + \mathcal{L}_{\Omega}$ &  139.3 \\ 
$\mathcal{L}_{D} + \mathcal{L}_{\Omega} + \mathcal{L}_{base} $ & 122.1 \\ 
$\mathcal{L}_{D} + \mathcal{L}_{\Omega} +  \mathcal{L}_{base}  + \mathcal{L}_{NF}$ & 112.3 \\ 
\hline
$\mathcal{L}_{D} + \mathcal{L}_{\Omega} + \mathcal{L}_{base}  + \mathcal{L}_{NF} + \mathcal{L}_{bl} $      & \textbf{96.7} \\
\hline
\end{tabular}
\end{center}
\caption{\textbf{Ablation studies.} Experiments with different loss terms using Human3.6M dataset for training and testing.}
\label{table:ablation-table}
\end{table}

\section{Discussion}
Our method outperforms self-supervised state-of-the-art approaches that estimate 3D pose from images and assume unpaired 3D data for supervision \cite{kundu2020self}. Also, it performs better than recent methods that rely on 3D supervision \cite{chen20173d} or multi-view images \cite{rhodin2018learning, rhodin2018unsupervised}. Moreover, its performance is similar to one method that assume 3D kinematic constraints \cite{kundu2020kinematic}. We achieve superior performance to Kundu \etal \cite{kundu2020kinematic} in $20\%$ of the activities (\textit{Discuss, Pose,} and \textit{Wait}), and close scores for the rest.

\begin{figure}[ht]
\centering
\includegraphics[width=\linewidth]{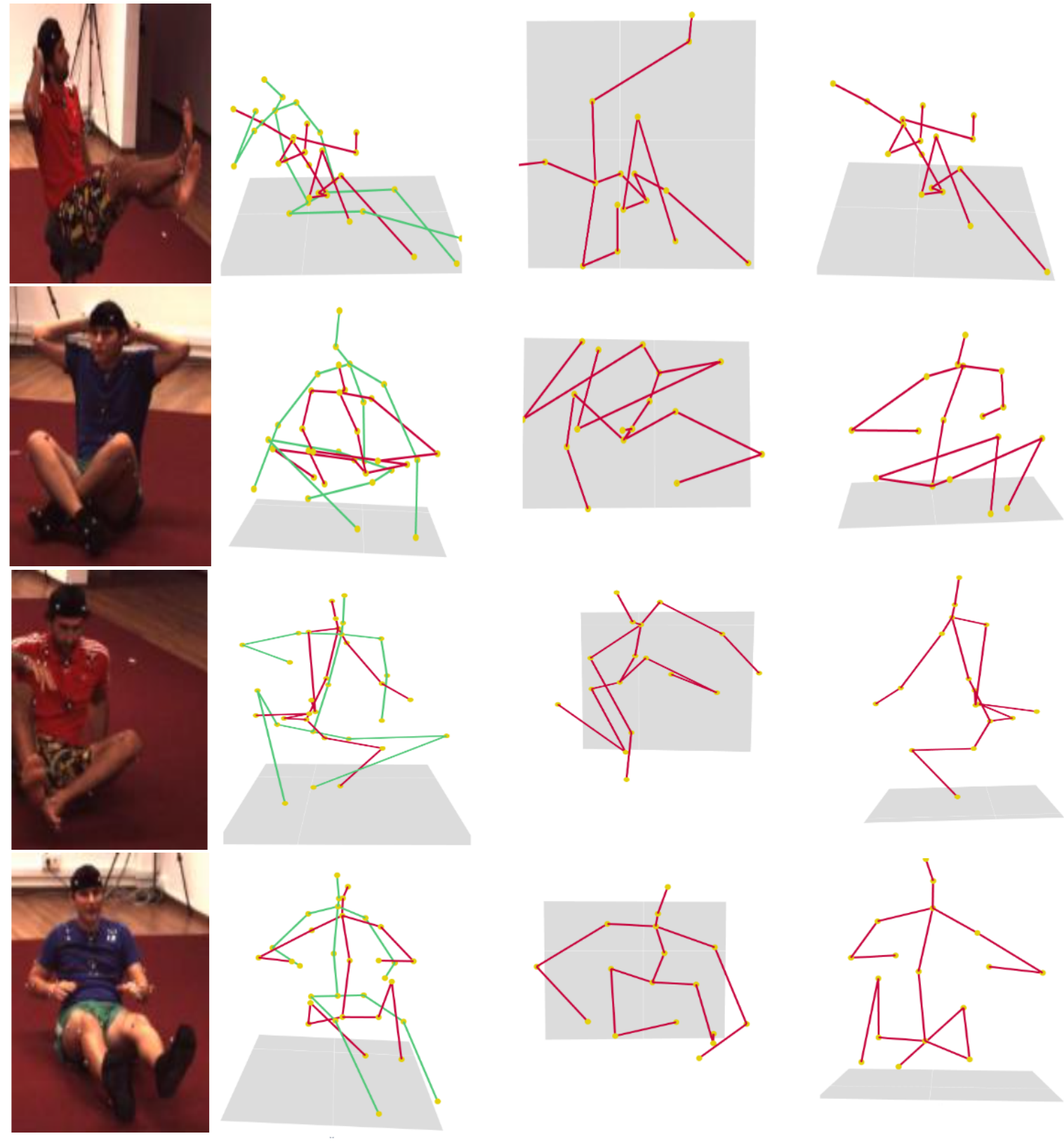}
\caption{\textbf{Failure cases on Human3.6M.} 3D predictions with a P-MPJPE greater than $200$mm. The first column shows the input images. The second column displays the predicted 3D pose (coloured in red) aligned with its respective ground truth (coloured in green). Following columns show different views of the predicted 3D pose.} 
\label{fig:vis-fil}
\end{figure}

Most failure cases of our model on the Human3.6M dataset appear for snapshots from activities such as \textit{Sitting} and \textit{Sitting Down}. We assume this occurs because of the self-occlusions and perspective ambiguity in these activities. However, according to the examples shown in \autoref{fig:vis-fil}, the model can still produce plausible 3D poses for most cases, even if they do not exactly match their respective 3D ground truth. The high P-MPJPE comes from mismatches between the `joints' representing the body's extremities, e.g. hands and feet.

With the experiments using the MPI-INF-3DHP dataset, we demonstrate the cross-dataset generalisation of our method. We also show that it works well even if we train the NF network with a different dataset (Human3.6M). Moreover, when training with images from Human3.6M and MPI-INF-3DHP, the experimental results suggest that increasing the diversity of images in the training set could help to improve the overall performance. In principle, extending the dataset of images is relatively straightforward since 3D annotations are not require.

\section{Conclusion}
We demonstrate how to estimate 3D human pose with a training architecture requiring only images depicting people in different poses and an unpaired set of typical 2D poses. We demonstrate qualitatively that our approach holds the potential for rapidly learning about the pose of articulated structures other than the human body without the need to collect ground-truth 3D pose data, e.g. human hands. 

Overall, using human datasets the qualitative and quantitative results suggest that our method is comparable to other self-supervised state-of-the-art approaches that estimate 3D pose from images despite requiring a less onerous dataset for training. Furthermore, it performs better than state-of-the-art methods that rely on multi-view images or 3D pose annotations for supervision. Prior work has demonstrated the value of using temporal information from image sequences and domain adaptation networks. Incorporating these into our approach would be a promising direction for future work. Finally, the way is open to apply the method to much larger datasets of unlabelled images to see whether performance continues to improve, and to apply the method to other articulated structures (e.g., mice, dogs and other animals), exploiting the relatively light requirement for self-supervision in the form of an empirical prior on 2D poses.

\section*{Acknowledgments} The first author is a recipient of a CONACYT Scholarship. Special thanks to Rebecca Stone and Mohammed Alghamdi for great discussions and feedback.

%%%%%%%%% REFERENCES
{\small
\bibliographystyle{ieee_fullname}
\bibliography{egbib}
}

\end{document}